\documentclass[12pt]{article}
\usepackage{scicite}
\usepackage{times}
\usepackage{url}
\usepackage{xcolor}
\usepackage{graphicx}
\usepackage{hyperref}
\usepackage{cellspace}

\newcommand\cincludegraphics[2][]{\raisebox{+0.4\height}{\includegraphics[#1]{#2}}}

\newenvironment{sciabstract}{%
\begin{quote} \bf}
{\end{quote}}

\title{Generative and reproducible benchmarks \\for comprehensive evaluation \\of machine learning classifiers}

\author
{Patryk Orzechowski,$^{1,2\ast}$ Jason H. Moore,$^{1\ast}$

\\
\\
\normalsize{$^{1}$Institute for Biomedical Informatics, University of Pennsylvania,}\\
\normalsize{Philadelphia, PA 19104, USA }\\
\normalsize{$^{2}$Department of Automatics and Robotics, AGH University of Science and Technology}\\
\normalsize{al. Mickiewicza 30, 30-059 Krakow, Poland} \\
\normalsize{$^\ast$To whom correspondence should be addressed:}\\
\normalsize{E-mail:  patryk.orzechowski@gmail.com, jhmoore@upenn.edu.}
}

\date{}

\begin{document} 


\baselineskip 18pt


\maketitle


\begin{sciabstract}
Understanding the strengths and weaknesses of machine learning (ML) algorithms is crucial for determine their scope of application. Here, we introduce the DIverse and GENerative ML Benchmark  (DIGEN) -- a collection of synthetic datasets  for comprehensive, reproducible, and interpretable benchmarking of machine learning algorithms for classification of binary outcomes. The DIGEN resource consists of 40 mathematical functions which map continuous features to discrete endpoints for creating synthetic datasets. These 40 functions were discovered using a heuristic algorithm designed to maximize the diversity of performance among multiple popular machine learning algorithms thus providing a useful test suite for evaluating and comparing new methods. Access to the generative functions facilitates understanding of why a method performs poorly compared to other algorithms thus providing ideas for improvement. The resource with extensive documentation and analyses is open-source and available on GitHub.
\end{sciabstract}

\section*{Introduction}
The development of new machine learning algorithms has accelerated to meet the demands of a variety of big data applications. An important type of machine learning algorithm is the classifier which is designed to accept discrete and/or continuous input features and produce a binary prediction or outcome which matches as close as possible a binary target such as presence or absence of disease or success or failure of a device. This class of algorithms, sometimes referred to as supervised machine learning, is useful in many domains and is often used to complement parametric statistical methods such as logistic regression. Examples include tree-based methods such as decision trees \cite{breiman1984classification}, random forests \cite{Breiman2001}, kernel-based methods such as support vector machines \cite{cortes1995support}, and gradient boosted trees \cite{Friedman2001}, as well as and their many variants \cite{Chen2016,ke2017lightgbm}.

Central to the development of machine learning algorithms is their evaluation. A good evaluation should document the strengths and weaknesses of the method and allow a fair and robust comparison to other state-of-the-art methods. Evaluation criteria often include measures of the accuracy of the predictions made, the computational efficiency of the algorithms, the degree of fairness and bias, and the user’s ability to interpret the results from a fitted model. Evaluation results help practitioners understand when it is appropriate to use the method, help readers assess the trustworthiness of the results, and help developers generate ideas for how to make improvements. One approach to evaluation is to use real and/or simulated datasets as ‘benchmarks’. Two commonly used collections of real and synthetic data are the University of California Irvine (UCI) machine learning repository \cite{Dua:2019}, and the Library for Support Vector Machines (LIBSVM) \cite{chang2011libsvm} which provide hundreds real datasets with open access. Public efforts focused on reorganizing, standardizing, and expanding those repositories lead to emergence of benchmarking repositories, such as the Penn Machine Learning Benchmark (PMLB) \cite{olson2017pmlb}, Open Machine Learning 100 (OpenML100) and its curated successor OpenML-CC18 \cite{bischl2017openml}. Lot of effort has been made in order to organize benchmarks for  regression problems \cite{olson2017pmlb, orzechowski2018we, udrescu2020ai, la2021contemporary}. Benchmarks have also become more popular by competitions such as Kaggle \cite{network2011kaggle} that provide data for the comparison of algorithms provided by contestants. The results of such competitions have helped the machine learning community evaluate and improve numerous algorithms. 

Increased data accessibility allowed extensive testing of different ML algorithms. A common benchmarking practice is to select a subset of these datasets to illustrate one method performing better than others. There are several problems with this approach. Most importantly, it is rare to know what the true patterns are in real data. As such, it is difficult know whether an algorithm is performing well because it is modeling the truth or exploiting the noise in the data (i.e. overfitting). This issue can be addressed by the inclusion of replicate datasets drawn from the same experiment or observational study. Ideally, at least three datasets of sufficient sample size and consistent structure would be available for training a model (dataset one), tuning the parameters of the model (dataset two), and validation (dataset three). Additional validation datasets provide additional confidence that the model is generalizable and thus likely to be modeling the signal in the data. However, real data are time consuming and expensive and thus multiple real datasets are rarely available. Even when multiple datasets are available, it is often difficult to know whether the examples are representative of the population they were drawn from or whether the features were measured in the same way. Further, the release of real data to the public can be problematic because of privacy, intellectual property, or confidentiality issues. These limitations have led some to turn to simulated data. 

A major advantage of simulated data is that the ground truth is known because the signal and noise are specifically engineered. This may provide some insights into why a method perform better or worse on certain datasets. Further, a generative function makes it possible for the user to generate as many replicate datasets and with any shape to evaluate the methods. Of course, an important limitation of simulated data is that it may not be possible to know whether the patterns being generated are consistent with those from real data. Despite the strengths, there is a noticeable lack of simulated data for benchmarking classification algorithms. Additional limitation is that existing datasets are not focused on differentiating the accuracy of the classifiers. Thus, multiple methods commonly end up having similar performance for multiple available datasets. Those datasets in terms of benchmarking are non-informative and abundant as they do not highlight sufficiently strengths and weaknesses of classification methods. 

There are several aspects that a good synthetic benchmark should deliver. A desirable quality of a good synthetic benchmark is not only the ability to differentiate the performance of multiple methods within a dataset, but also between datasets, such that no one method would dominate all the others (this is commonly referred as \emph{no free lunch theorem} in machine learning). A useful benchmark should also deliver insights regarding types of generative functions that are not suitable for the method being evaluated so that strategies for improving the method can be developed. It should be compact to offer rapid evaluation, yet comprehensive to cover varying types of problems. Ideally, it would cover small problems, but would also provide possibility to scale them to larger problems of any size, with the same built-in ground truth. It would be also great if the benchmark provided pre-computed information on expected performance of the methods on replicated datasets initialized with different random seeds. Finally, the benchmark should be reproducible, so that the results across multiple machines and operating systems remained the same.

With aforementioned qualities in mind, we introduce the \textbf{DIverse and GENerative (DIGEN)}, a synthetic data resource for comprehensive, reproducible, and interpretable benchmarking of machine learning algorithms for classification of binary outcomes. A central goal is to generate a diverse set of mathematical functions which map continuously distributed features to binary outcomes or class variables for the purpose of revealing the strengths and weaknesses of the machine learning algorithms being evaluated.

\section*{Methods}

The resulting collection of datasets has been created as a result of multi-step optimization built on top of a heuristic algorithm that discovers the generative mathematical functions. The datasets are simulated in a specific way that yields maximum diversity of the performance of multiple commonly used machine learning algorithms. The heuristic algorithm has two optimization objectives: accuracy (measured as AUROC between two selected methods), and standard deviation between the remaining  methods. The candidate functions have emerged as the result of duels between pairs of machine learning algorithms from the following list: decision trees, gradient boosting, k-nearest neighbors \cite{cover1967nearest}, light gradient boosting (LightGBM) \cite{ke2017lightgbm}, logistic regression \cite{mccullagh1984generalized}, random forests, support vector machines, and extreme gradient boosting (XGBoost) \cite{Chen2016}. Within each duel, the endpoint of the datasets was modified so that one method excelled and the other underperformed. Additional requirement was to maximize diversity of the performance of all the algorithms. Using the optimization objectives, we selected 40 benchmarks with unique generative mathematical functions maximizing the diversity and ranking of machine learning algorithm performance. We provide not only synthetic datasets, but also the generative functions used as the ground truth, multiple different analyses for each of the datasets, source code for running analysis and a Docker script to replicate our study - everything as an open-source contribution for the machine learning community. 

Details of the heuristic algorithm and its implementation can be found in the Supplementary material, which is available on GitHub Pages. Briefly, the DIGEN resource was built using a heavily modified version of the Heuristic Identification of Biological Architectures for Simulating Complex Hierarchical Genetic Interactions (HIBACHI) method and software \cite{moore2018computational} reinforced by selection of Pareto-optimal solutions using NSGA-III strategy \cite{deb2013evolutionary} and the state-of-the-art hyper-parameters optimization framework Optuna \cite{optuna_2019}. The popular scikit-learn machine learning library \footnote{https://pypi.org/project/scikit-learn/} \cite{scikit-learn} was used for analysis with addition of xgboost \footnote{https://pypi.org/project/xgboost/} and lightgbm \footnote{https://pypi.org/project/lightgbm/} Python packages. Parameter ranges were set based on the recommendations of the leading hyper-optimization frameworks, such as Optuna, auto-sklearn \cite{ASKL,ASKL2}, hyperopt \cite{bergstra2013making}, and expert knowledge. The training and testing datasets were split 80/20, we used 10-fold cross-validation. The exact parameter settings of the methods could be found on GitHub.  

The DIGEN benchmark resource includes 40 datasets simulated from each of the generative mathematical functions. Each includes 10 features or independent variable generated from a normal distribution with a mean of zero and a variance of one - $N(0,1)$. The generative mathematical function accepts these features as input returning a continuously distributed outcome variable. We then sort and convert this outcome to binary values $(0,1)$ in order to create a balanced binary class variable. Each of the 40 datasets has a sample size of $n=1000$. We provide each of these 40 datasets as the benchmark.  
Datasets initiated with different random seeds, or with different sizes, could be generated using a Docker container available at GitHub. The functions themselves can be used with features drawn from the same normal distribution to generate as many datasets as desired with any sample and feature sizes. It would not be difficult to extend the heuristic algorithm to generate functions mapping other distributions of input and output data for problems such as those with binary inputs or multi-class or continuous outcomes.

\section*{Results}

\begin{figure}[htb]
\centering
  \begin{tabular}{cc}
   \includegraphics[width=.5\textwidth]{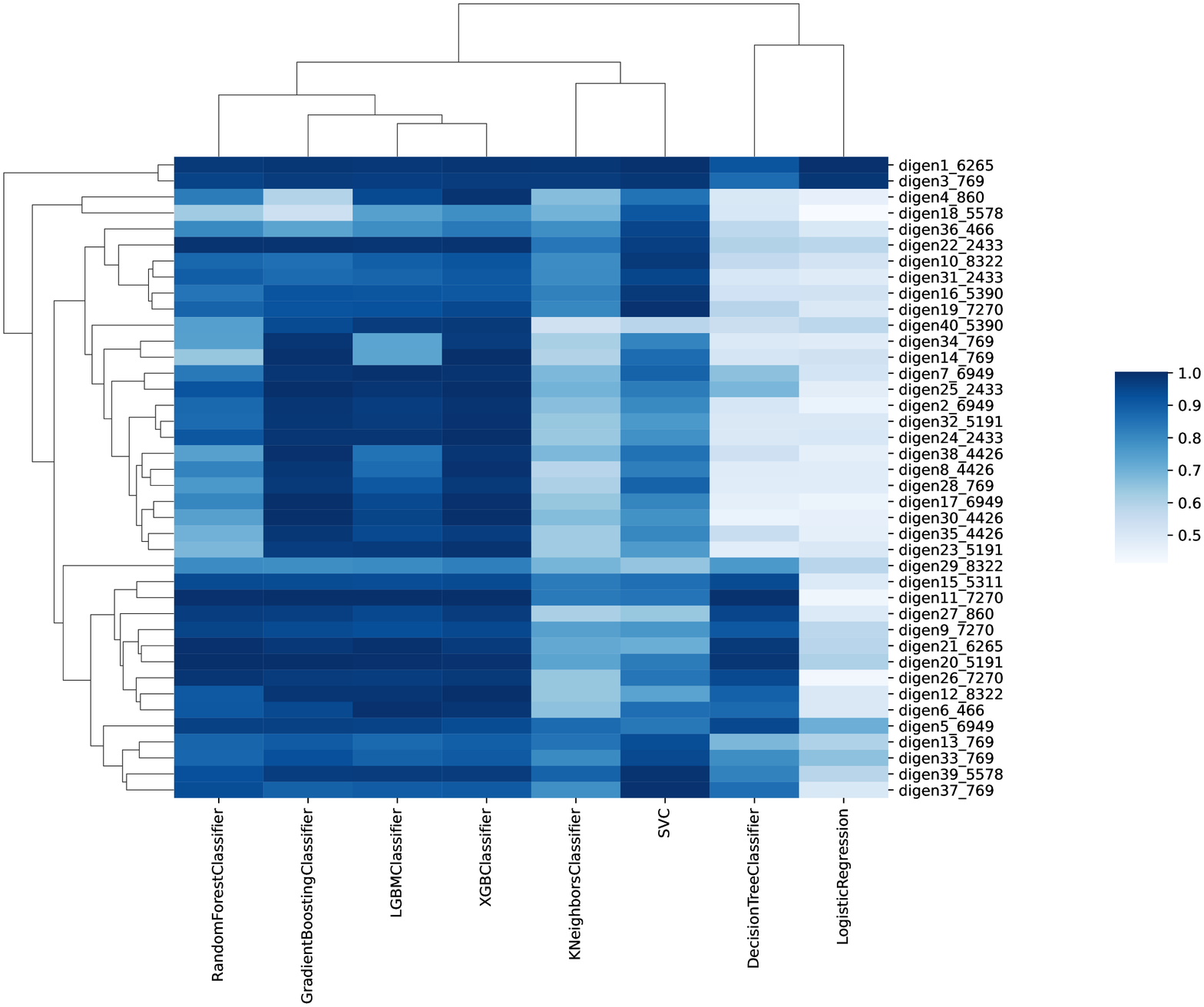} &
  \includegraphics[width=.5\textwidth]{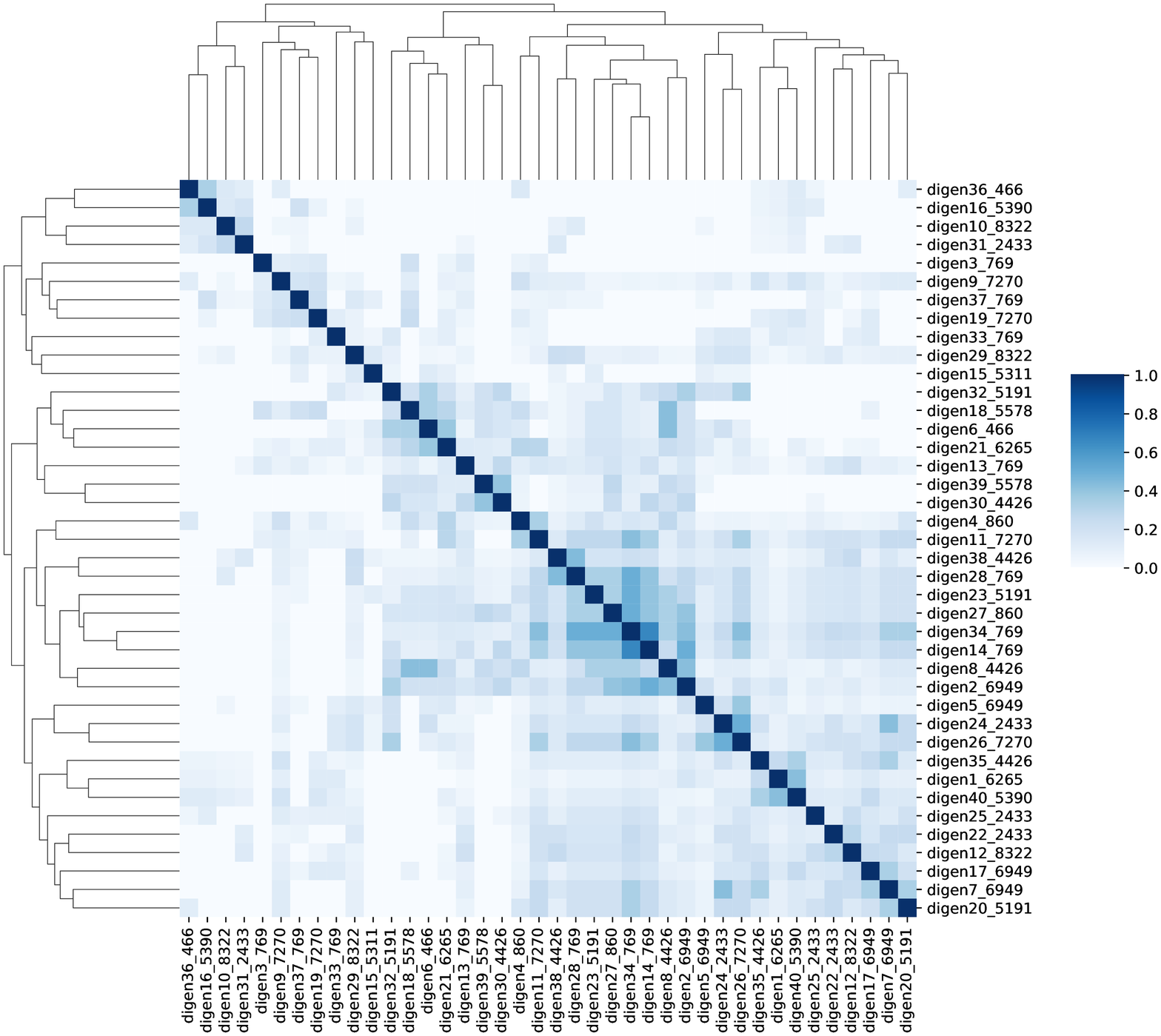} \\
A.&B.\\ 

   \\
  
  \end{tabular}
  \caption{Performance and similarity of machine learning methods across the 40 DIGEN benchmark datasets. \textbf{A}: Heatmap of AUROC scores for each machine learning method (columns) with respect to each of the datasets (rows).  \textbf{B.} Heatmap of Ruzicka similarity \cite{ruzicka1958anwendung} between each pair of datasets with regard to the number of common mathematical operators executed in proper order in the generative functions. All machine learning methods were tuned with respect to their hyper-parameters.}
  \label{benchmark}
\end{figure}

\begin{figure}[htb!]
\centering
  \begin{tabular}{@{}cc@{}}
    \includegraphics[width=.4\textwidth]{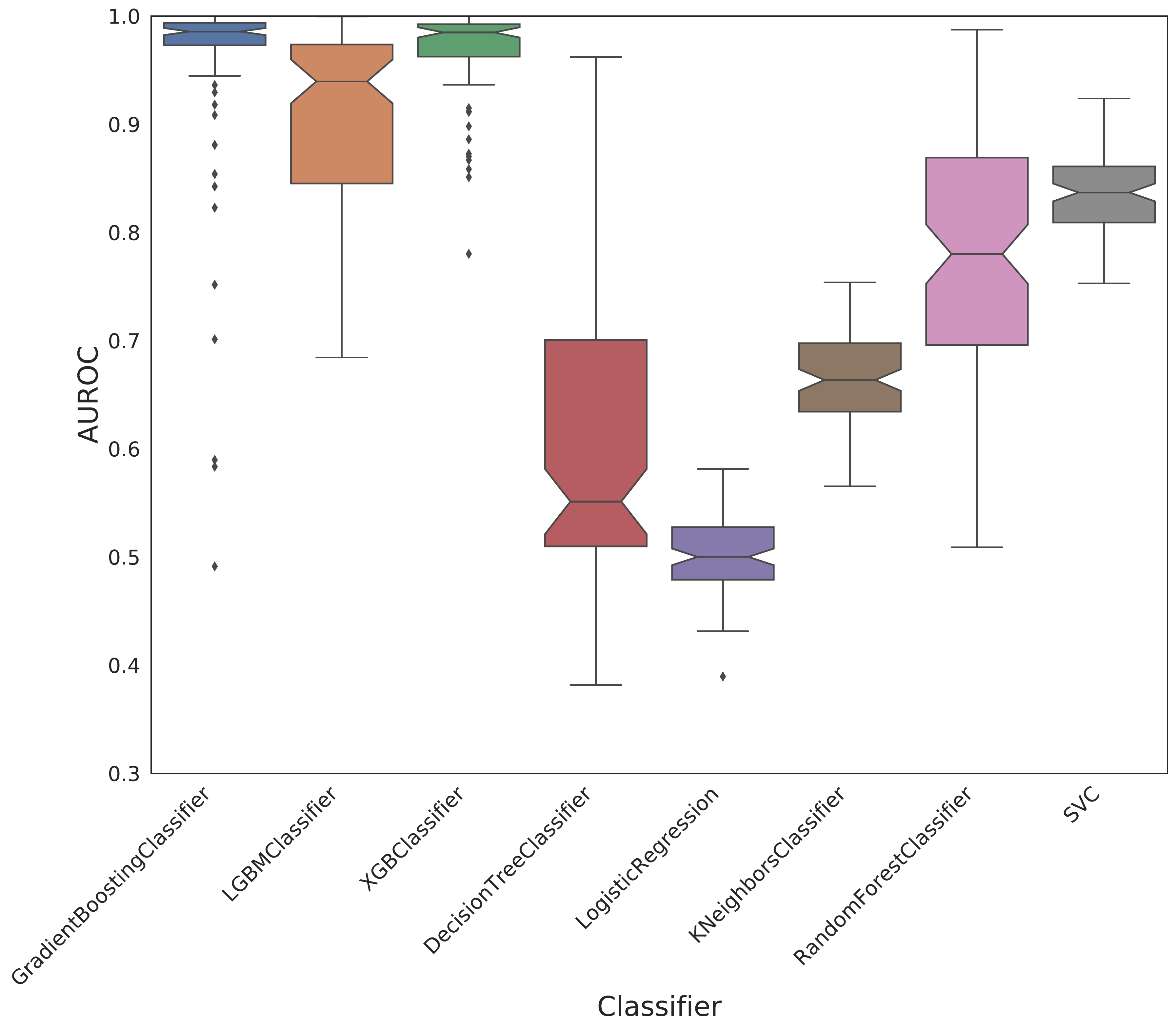} &
    \cincludegraphics[width=.4\textwidth]{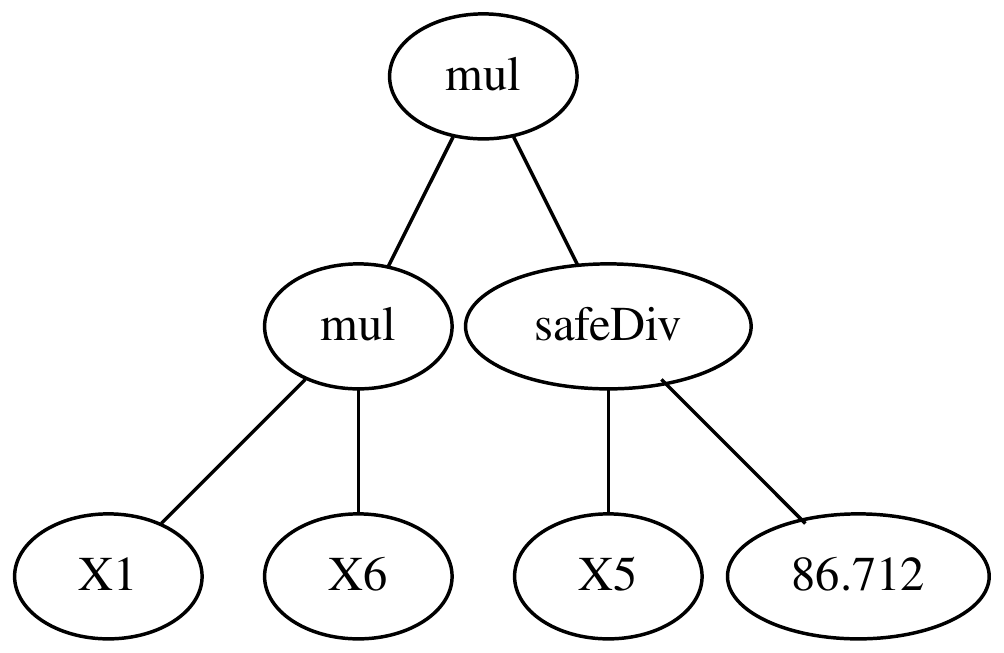}\\
    A.&B.\\  
    \includegraphics[width=.5\textwidth]{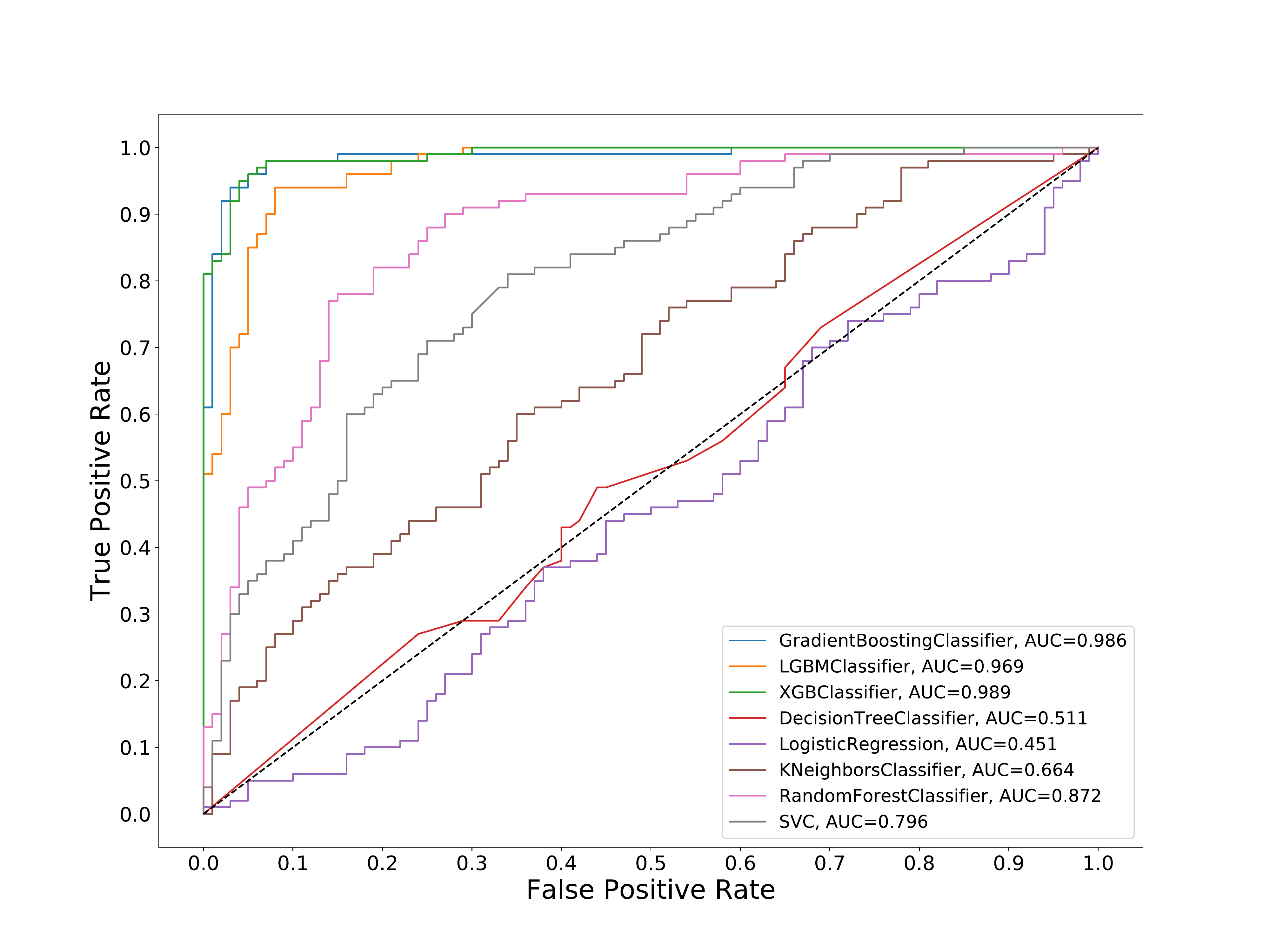}&
    \includegraphics[width=.5\textwidth]{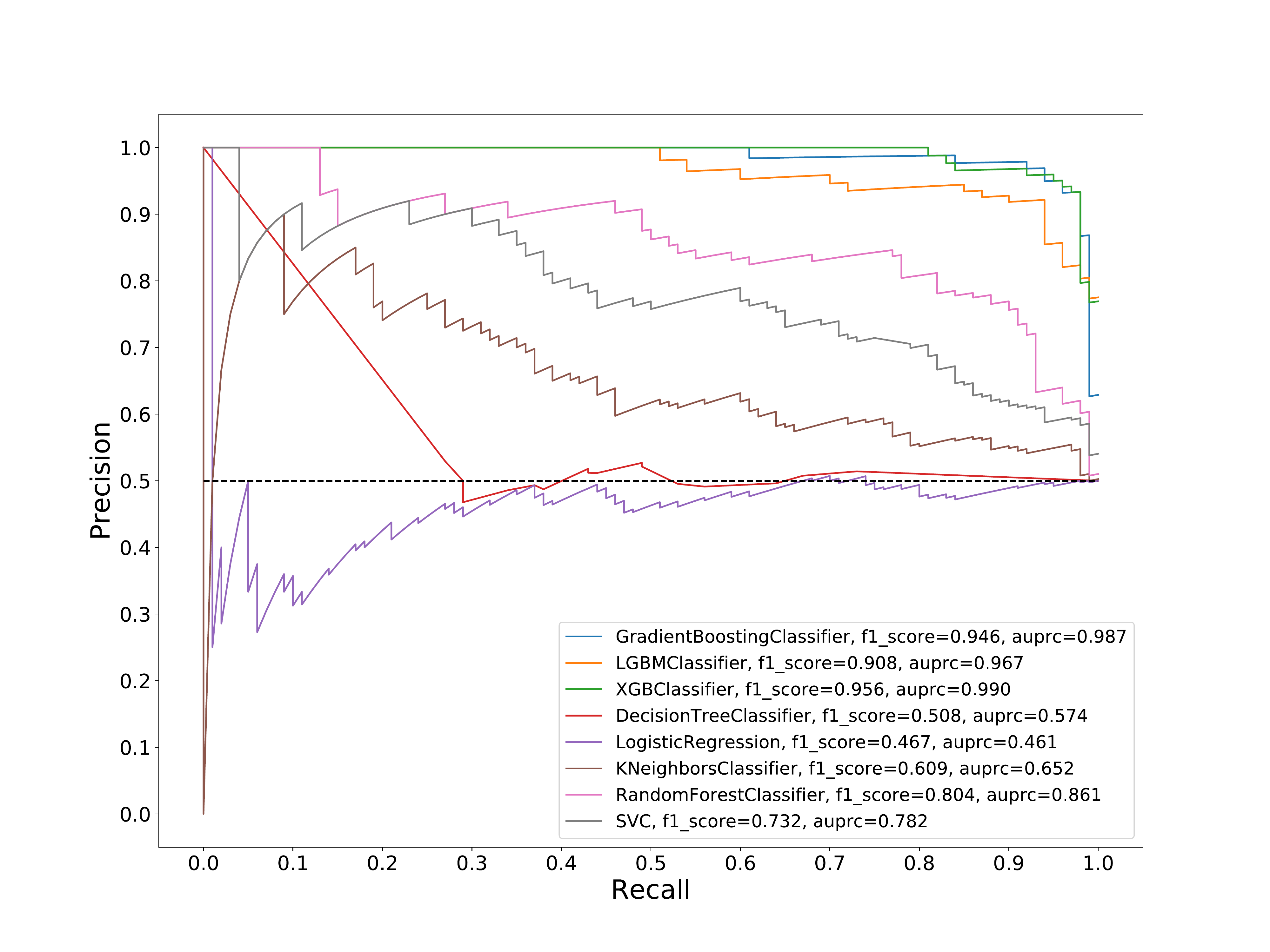}\\

    C.&D.\\
  \end{tabular}
  \caption{\textbf{A.} Boxplot showing performance of each tuned machine learning method after 100 iterations of hyper-parameter tuning on 100 replicate datasets initialized with different random seeds. \textbf{B.} The mathematical model used to generate the discrete outcome for this dataet (safeDiv performs a safe division which prevents dividing by zero). \textbf{C.} Plot showing combined ROC curves for machine learning methods run on the dataset initiated with the default seed. \textbf{D.} Plot showing combined PRC plots for machine learning methods.} 
  
  \label{digen40_charts}
\end{figure}

There are several properties which make DIGEN a unique benchmark in the machine learning community. First, DIGEN \emph{differentiates} the performance of multiple machine learning methods. Figure \ref{benchmark}A shows a heatmap of the performance of reference machine learning algorithms (presented in columns) initialized with the default random seed and optimized with 200 hyper-parameter evaluations across all 40 benchmark datasets (rows). The shade of the color is proportional to the area under the receiver operating characteristic (AUROC). Also shown is the results of a hierarchical cluster analysis of the rows and columns visualized using dendrograms with shorter branches indicating higher similarity. Note that no method dominates all datasets. Second, DIGEN includes a \emph{diverse} set of generative mathematical functions. Figure \ref{benchmark}B shows a heatmap of the Ruzicka similarity of the 40 generative functions. The Ruzicka similarity between each pair of datasets represented as vectors $x$ and $y$ is calculated as the number of mathematical pairs of operators used one after another in the generative function, and is defined as $\sum \min{(x_i,y_i)} \over \sum \max{(x_i, y_i)}$. Note the low similarity between all benchmark functions. Third, DIGEN is \emph{comprehensive}. The heuristic algorithm optimization considered hundred of thousands of combinations of mathematical operators using millions of hours of computing time. Further, every machine learning algorithm considered by the heuristic algorithm was tuned by with 200 hyper-parameter combinations evaluated based on AUROC score computed with 10-fold cross-validation. 

In addition to the computational properties outlined above, DIGEN has several practical qualities for benchmarking. First, DIGEN datasets and results are \emph{reproducible}. We have provided a Docker container which allows to reproduce our analyses. New datasets can be generated from the same feature distributions and mathematical functions producing the same patterns. This will allow the benchmark to be recreated as necessary yielding similar machine learning results depending on the random seed used. Second, DIGEN is \emph{scalable}. The generative functions allow users to generate as many datasets as they want and with any desired sample size. DIGEN is also  \emph{compact}. The 40 benchmarks were selected from all of those generated by the heuristic to be diverse and to avoid unnecessary redundancy. The goal was to maximize the utility without burdening the user with too many benchmarks to evaluate. Third, DIGEN is \emph{simple}, as it allows the user to very quickly discover where and why their method is not performing as good as the reference methods. To facilitate this, we have provided the code in Jupyter Notebooks for running the analyses and comparing the machine learning methods covered in the benchmark. We have included multiple pre-computed statistics for each of the 40 generative functions, such as a feature correlation chart, boxplots (Fig. \ref{digen40_charts}A) which reflect the optimized performance with 100 evaluations of the machine learning algorithms across 100 replicate datasets initiated with different random seeds (Fig. \ref{digen40_charts}B), ROC plots (Fig. \ref{digen40_charts}C) and precision-recall curve (PRC) plots  (Fig. \ref{digen40_charts}D). All the statistics were computed using tuned machine learning methods with a specific random seed given next to the name of the DIGEN dataset. Finally, DIGEN is  \emph{open source}. All the source code along with the benchmark datasets are available on GitHub.


\begin{figure}[htb]
\centering
  \begin{tabular}{@{}cc@{}}

\includegraphics[width=.9\textwidth]{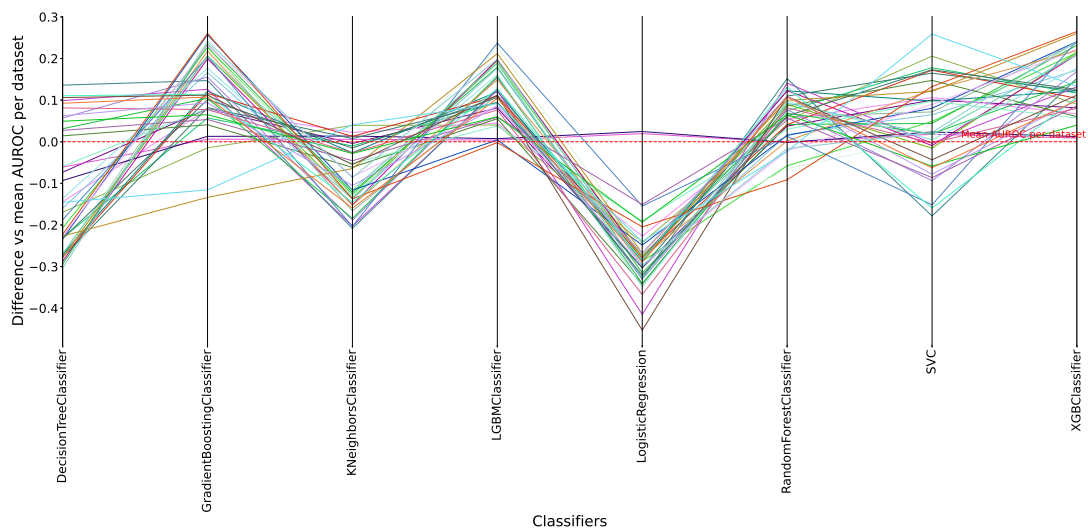}    \\
  \end{tabular}
  \caption{Comparative performance of different ML models on DIGEN. Positive values indicate that the analyzed ML method had better accuracy than the mean accuracy across all of the methods.}
    
  \label{digen:comparison}
\end{figure}

\section*{Discussion}

Despite the widespread use of benchmarks, there is general recognition that they have important limitations both in their design and their application. We review a few of those limitations here and then present the results of a study designed to generate a comprehensive suite of simulated benchmarks designed to reveal the strengths of weaknesses of machine learning classification methods.

In summary, we have generated a comprehensive set of synthetic benchmark data to facilitate the evaluation and comparison of machine learning algorithms for classification of discrete outcomes. Each benchmark dataset comes with the generative mathematical function which can be used to create additional datasets or be used to provide some clues as to why a machine learning algorithm might not be performing well on that particular generative function. The resource is diverse, easy to use, and open-source. Further, it includes tutorials, the source code and  mathematical formulas used to simulate the data. This allows to easily extend the resource or benchmark the methods under similar environment. As emphasized by Sculley et al. \cite{Sculley2018WinnersCO}, DIGEN is not only a tool to count the wins of a particular machine learning method, but is intended to be a helpful tool for explaining \emph{where} the method under performs and hopefully also \emph{why}.

\bibliography{scibib}

\begin{thebibliography}{10}

\bibitem{optuna_2019}
Takuya Akiba, Shotaro Sano, Toshihiko Yanase, Takeru Ohta, and Masanori Koyama.
\newblock Optuna: A next-generation hyperparameter optimization framework.
\newblock In {\em Proceedings of the 25rd {ACM} {SIGKDD} International
  Conference on Knowledge Discovery and Data Mining}, 2019.

\bibitem{bergstra2013making}
James Bergstra, Daniel Yamins, and David Cox.
\newblock Making a science of model search: Hyperparameter optimization in
  hundreds of dimensions for vision architectures.
\newblock In {\em International conference on machine learning}, pages
  115--123, 2013.

\bibitem{bischl2017openml}
Bernd Bischl, Giuseppe Casalicchio, Matthias Feurer, Frank Hutter, Michel Lang,
  Rafael~G Mantovani, Jan~N van Rijn, and Joaquin Vanschoren.
\newblock Openml benchmarking suites and the openml100.
\newblock {\em arXiv preprint arXiv:1708.03731}, 2017.

\bibitem{Breiman2001}
Leo Breiman.
\newblock Random forests.
\newblock {\em Machine learning}, 45(1):5--32, 2001.

\bibitem{breiman1984classification}
Leo Breiman, Jerome Friedman, Charles~J Stone, and Richard~A Olshen.
\newblock {\em Classification and regression trees}.
\newblock CRC press, 1984.

\bibitem{chang2011libsvm}
Chih-Chung Chang and Chih-Jen Lin.
\newblock Libsvm: A library for support vector machines.
\newblock {\em ACM transactions on intelligent systems and technology (TIST)},
  2(3):1--27, 2011.

\bibitem{Chen2016}
Tianqi Chen and Carlos Guestrin.
\newblock Xgboost: A scalable tree boosting system.
\newblock In {\em Proceedings of the 22nd acm sigkdd international conference
  on knowledge discovery and data mining}, pages 785--794. ACM, 2016.

\bibitem{cortes1995support}
Corinna Cortes and Vladimir Vapnik.
\newblock Support-vector networks.
\newblock {\em Machine learning}, 20(3):273--297, 1995.

\bibitem{cover1967nearest}
Thomas Cover and Peter Hart.
\newblock Nearest neighbor pattern classification.
\newblock {\em IEEE transactions on information theory}, 13(1):21--27, 1967.

\bibitem{deb2013evolutionary}
Kalyanmoy Deb and Himanshu Jain.
\newblock An evolutionary many-objective optimization algorithm using
  reference-point-based nondominated sorting approach, part i: solving problems
  with box constraints.
\newblock {\em IEEE transactions on evolutionary computation}, 18(4):577--601,
  2013.

\bibitem{Dua:2019}
Dheeru Dua and Casey Graff.
\newblock {UCI} machine learning repository, 2017.

\bibitem{ASKL2}
Matthias Feurer, Katharina Eggensperger, Stefan Falkner, Marius Lindauer, and
  Frank Hutter.
\newblock Auto-sklearn 2.0: The next generation.
\newblock {\em arXiv preprint arXiv:2007.04074}, 2020.

\bibitem{ASKL}
Matthias Feurer, Aaron Klein, Katharina Eggensperger, Jost Springenberg, Manuel
  Blum, and Frank Hutter.
\newblock Efficient and robust automated machine learning.
\newblock In C.~Cortes, N.~D. Lawrence, D.~D. Lee, M.~Sugiyama, and R.~Garnett,
  editors, {\em Advances in Neural Information Processing Systems 28}, pages
  2962--2970. Curran Associates, Inc., 2015.

\bibitem{Friedman2001}
Jerome~H Friedman.
\newblock Greedy function approximation: a gradient boosting machine.
\newblock {\em Annals of statistics}, pages 1189--1232, 2001.

\bibitem{ke2017lightgbm}
Guolin Ke, Qi~Meng, Thomas Finley, Taifeng Wang, Wei Chen, Weidong Ma, Qiwei
  Ye, and Tie-Yan Liu.
\newblock Lightgbm: A highly efficient gradient boosting decision tree.
\newblock In {\em Advances in neural information processing systems}, pages
  3146--3154, 2017.

\bibitem{la2021contemporary}
William La~Cava, Patryk Orzechowski, Bogdan Burlacu, Fabr{\'\i}cio~Olivetti
  de~Fran{\c{c}}a, Marco Virgolin, JIN Ying, Michael Kommenda, and Jason~H
  Moore.
\newblock Contemporary symbolic regression methods and their relative
  performance.
\newblock 2021.

\bibitem{mccullagh1984generalized}
Peter McCullagh.
\newblock Generalized linear models.
\newblock {\em European Journal of Operational Research}, 16(3):285--292, 1984.

\bibitem{moore2018computational}
Jason~H Moore, Randal~S Olson, Peter Schmitt, Yong Chen, and Elisabetta
  Manduchi.
\newblock How computational thought experiments can improve our understanding
  of the genetic architecture of common human diseases.
\newblock In {\em Artificial Life Conference Proceedings}, pages 23--30. MIT
  Press, 2018.

\bibitem{network2011kaggle}
Kaggle Network.
\newblock Kaggle network: Your home for data science, 2011.

\bibitem{olson2017pmlb}
Randal~S Olson, William La~Cava, Patryk Orzechowski, Ryan~J Urbanowicz, and
  Jason~H Moore.
\newblock Pmlb: a large benchmark suite for machine learning evaluation and
  comparison.
\newblock {\em BioData mining}, 10(1):1--13, 2017.

\bibitem{orzechowski2018we}
Patryk Orzechowski, William La~Cava, and Jason~H Moore.
\newblock Where are we now? a large benchmark study of recent symbolic
  regression methods.
\newblock In {\em Proceedings of the Genetic and Evolutionary Computation
  Conference}, pages 1183--1190, 2018.

\bibitem{scikit-learn}
F.~Pedregosa, G.~Varoquaux, A.~Gramfort, V.~Michel, B.~Thirion, O.~Grisel,
  M.~Blondel, P.~Prettenhofer, R.~Weiss, V.~Dubourg, J.~Vanderplas, A.~Passos,
  D.~Cournapeau, M.~Brucher, M.~Perrot, and E.~Duchesnay.
\newblock Scikit-learn: Machine learning in {P}ython.
\newblock {\em Journal of Machine Learning Research}, 12:2825--2830, 2011.

\bibitem{ruzicka1958anwendung}
M~Ruzicka.
\newblock Anwendung mathematisch-statisticher methoden in der geobotanik
  (synthetische bearbeitung von aufnahmen).
\newblock {\em Biologia, Bratisl}, 13:647--661, 1958.

\bibitem{Sculley2018WinnersCO}
D.~Sculley, Jasper Snoek, Alexander~B. Wiltschko, and Ali Rahimi.
\newblock Winner's curse? on pace, progress, and empirical rigor.
\newblock In {\em ICLR}, 2018.

\bibitem{udrescu2020ai}
Silviu-Marian Udrescu and Max Tegmark.
\newblock Ai feynman: A physics-inspired method for symbolic regression.
\newblock {\em Science Advances}, 6(16):eaay2631, 2020.

\end{thebibliography}
\bibliographystyle{plain}

\section*{Acknowledgments}
This research was supported in part by PL-Grid infrastructure and USA National Institutes of Health (NIH) grants AI116794, LM010098, and LM012601. The authors would like thank Dr. William La Cava for his helpful comments and useful discussion.

\section*{Supplementary materials}
All the source code along with the tutorials and extensive documentations are open source and available at \url{https://github.com/EpistasisLab/digen/}. Living version of Supplementary material with description of the methods is available at \url{https://epistasislab.github.io/digen/}.

\section*{Abbreviations}

\begin{description}
\item AUROC - area under receiver operating characteristics curve
\item ROC - receiver operating characteristics
\item PRC - precission-recall curve
\item ML - Machine Learning
\end{description}


    








\end{document}